\documentclass[sigconf=false,screen,review=false, anonymous=false, nonacm=true]{acmart}
\AtBeginDocument{%
  \providecommand\BibTeX{{%
    \normalfont B\kern-0.5em{\scshape i\kern-0.25em b}\kern-0.8em\TeX}}}

\usepackage{lipsum}
\usepackage[linesnumbered,vlined,ruled,commentsnumbered,algo2e]{algorithm2e}

\setcopyright{acmcopyright}
\copyrightyear{2023}
\acmYear{2023}
\acmDOI{XXXXXXX.XXXXXXX}

\acmConference[FOGA '23]{Foundations of Genetic Algorithms}{
  2023}{Potsdam, Germany}
\acmBooktitle{17th ACM/SIGEVO Conference on Foundations of Genetic Algorithms,
 August 30-- September 01, 2023, Potsdam, Germany} 

\usepackage{graphicx}
\usepackage{soul,xcolor}
\usepackage{amsmath}
\usepackage{subfigure}
\usepackage{hyperref}
\usepackage{csquotes}
\usepackage{algorithm}
\usepackage{algorithmicx}
\usepackage{algpseudocode}

\begin{document}

\title{Challenges of ELA-guided Function Evolution using Genetic Programming}

\author{Fu Xing Long}
\orcid{0000-0003-4550-5777}
\affiliation{%
  \institution{BMW Group}
  \streetaddress{Knorrstra{\ss}e 147}
  \city{Munich}
  \country{Germany}}
\email{fu-xing.long@bmw.de}

\author{Diederick Vermetten}
\orcid{0000-0003-3040-7162}
\affiliation{%
  \institution{LIACS, Leiden University}
  \streetaddress{Niels Bohrweg 1}
  \city{Leiden}
  \country{Netherlands}}
\email{d.l.vermetten@liacs.leidenuniv.nl}

\author{Anna V. Kononova}
\orcid{0000-0002-4138-7024}
\affiliation{%
  \institution{LIACS, Leiden University}
  \streetaddress{Niels Bohrweg 1}
  \city{Leiden}
  \country{Netherlands}}
\email{a.kononova@liacs.leidenuniv.nl}

\author{Roman Kalkreuth}
\orcid{0000-0003-1449-5131}
\affiliation{%
  \institution{LIACS, Leiden University}
  \streetaddress{Niels Bohrweg 1}
  \city{Leiden}
  \country{Netherlands}}
\email{r.t.kalkreuth@liacs.leidenuniv.nl}

\author{Kaifeng Yang}
\orcid{0000-0002-3353-3298}
\affiliation{%
  \institution{University of Applied Sciences \\Upper Austria}
  \streetaddress{Softwarepark 11, 4232}
  \city{Hagenberg}
  \country{Austria}}
\email{kaifeng.yang@fh-hagenberg.at}

\author{Thomas B\"{a}ck}
\orcid{0000-0001-6768-1478}
\affiliation{%
  \institution{LIACS, Leiden University}
  \streetaddress{Niels Bohrweg 1}
  \city{Leiden}
  \country{Netherlands}}
\email{t.h.w.baeck@liacs.leidenuniv.nl}

\author{Niki van Stein}
\orcid{0000-0002-0013-7969}
\affiliation{%
  \institution{LIACS, Leiden University}
  \streetaddress{Niels Bohrweg 1}
  \city{Leiden}
  \country{Netherlands}}
\email{n.van.stein@liacs.leidenuniv.nl}

\renewcommand{\shortauthors}{Fu Xing Long et al.}

\begin{abstract}
Within the optimization community, the question of how to generate new optimization problems has been gaining traction in recent years. 
Within topics such as instance space analysis (ISA), the generation of new problems can provide new benchmarks which are not yet explored in existing research. 
Beyond that, this function generation can also be exploited for solving complex real-world optimization problems. 
By generating functions with similar properties to the target problem, we can create a robust test set for algorithm selection and configuration.

However, the generation of functions with specific target properties remains challenging. 
While features exist to capture low-level landscape properties, they might not always capture the intended high-level features. 
We show that a genetic programming (GP) approach guided by these exploratory landscape analysis (ELA) properties is not always able to find satisfying functions. 
Our results suggest that careful considerations of the weighting of landscape properties, as well as the distance measure used, might be required to evolve functions that are sufficiently representative to the target landscape.
\end{abstract}

\ccsdesc[500]{Computing methodologies~Randomized search}



\keywords{function generator, genetic programming, exploratory landscape analysis, instance spaces}



\maketitle

\section{Introduction}

Benchmark problems play a key role in our ability to efficiently evaluate and compare the performance of optimization algorithms.
Well-constructed benchmark suites provide researchers the opportunity to gauge the different strengths and weaknesses of a wide variety of optimization algorithms~\cite{hansen2010comparing}.
Following this, carefully handcrafted sets of problems, such as the black-box optimization benchmarking (BBOB) suite~\cite{hansen2009bbob}, have become increasingly popular~\cite{bartz2020benchmarking}.
Beyond benchmarking purposes, the BBOB suite has been intensively used in the research field of algorithm selection problem (ASP)~\cite{rice1976algorithm}, with the focus on identifying computational and time-efficient algorithms for a particular problem instance.
Recently, this has been associated with the optimization landscape properties of problem instances, where the landscape properties are exploited to predict the performance of optimization algorithms, e.g., using machine learning models~\cite{pikalov2021automated,kerschke2019aas}.

One inherent limitation of these hand-crafted benchmark suites, however, lies in the fact that they can never cover the full instance space.
For instance, it has been shown that real-world automotive problem instances are insufficiently represented by the BBOB functions in terms of landscape properties~\cite{long2022learning,thomaser2022one}.
Consequently, there is a growing trend in understanding the coverage of problem classes by existing benchmark sets, and in the creation of new benchmarks to fill the gaps~\cite{smith2023instance, munoz2020generating}.
In this research area, or commonly known as instance space analysis (ISA), a feature-based representation of the problem instances are used to identify what functions are lacking and should be newly created.
This is often combined with a performance-oriented view of several optimization algorithms, leading to the creation of new functions, where the benefits of one algorithm over the others can clearly be observed.

The most common approach to generating new benchmark problems is through the use of genetic programming (GP)~\cite{munoz2020generating}. 
Since GP has a long history in the domain of symbolic regression, it is a natural choice for the creation of optimization problems.
Essentially, GP is guided towards a target feature vector in a poorly-covered part of the instance space.
These features can be generated, for example, using the exploratory landscape analysis (ELA)~\cite{mersmann2011exploratory}, which aims to capture low-level information about the problem landscape using a limited number of function evaluations. 

While this GP-based approach to function generation has shown considerable promise in ISA, it can also be used to create a set of surrogates for algorithm selection and hyperparameter tuning purposes.
This is especially useful for real-world optimization problems with expensive function evaluation, e.g., requiring simulator runs.
Indeed, previous work has shown that using benchmark problems with similar characteristics as surrogate for a real-world problem to tune optimization algorithms leads to performance benefits on the original function~\cite{thomaser2023transfer}.
Therefore, the ability to generate a set of problems with similar high-level problem features would be of significant practical importance.

In this work, we focus on investigating how a GP guided by ELA features can be utilized to generate problems which are similar to known benchmark functions. 
This illustrates the challenges which still need to be overcome to efficiently generate sets of feature-based surrogate problems. 
In particular, our contributions are as follows: 
\begin{itemize}
    \item We adapt the random function generator (RFG) from~\cite{tian2020arecommender} into a GP approach and investigate the impact this has on the distribution of ELA features of the generated problems.
    \item We investigate the impact of the used distance measure between ELA feature vectors. Our results show that equal treatment of all features might not be desirable.
\end{itemize}

The remainder of this paper is structured as follows:
Section~\ref{related_work} summarizes all the relevant topics for this work and Section~\ref{methodology} explains our methodology.
This is followed by the experimental setup in Section~\ref{experiment} and results are discussed in Section~\ref{results}.
Lastly, conclusions and future work are presented in Section~\ref{conclusion}.

\section{Related Work}\label{related_work}

\subsection{Black-Box Optimization Benchmarking}

The BBOB family of problem suites are some of the most well-known sets of problems for benchmarking optimization heuristic algorithms~\cite{hansen2010comparing}, particularly the original continuous, noiseless, single-objective suite, which is often referred to as \textit{the} BBOB~\cite{hansen2009bbob}.
To facilitate the benchmarking purposes, the BBOB suite has been integrated as part of the comparing continuous optimizers (COCO) platform~\cite{hansen2021coco} and iterative optimization heuristics profiler tool (IOHProfiler)~\cite{doerr2018ioh}, where the statistics of algorithm performances stored can be easily retrieved.
Due to its popularity, the BBOB suite has also become a common testbed for automated algorithm selection and configuration techniques, even though the suite was never designed with this in mind.

Altogether, the aforementioned original BBOB suite consists of $24$ functions from five problem classes based on their global properties.
While the BBOB functions were originally designed for unconstrained optimization, in practice, however, they are usually considered within the search domain of $[-5,5]^d$ with their global optimum located within $[-4,4]^d$.
Beyond the fact that they can be scaled to any arbitrary dimensionality $d$, the BBOB functions have the advantage that different variants or problem instances can be easily generated through a transformation of the search domain and objective values.
This transformation mechanism is internally integrated in BBOB and controlled by a unique identifier, or also called IID.

\subsection{Exploratory Landscape Analysis}
\label{subsec:ela}

In landscape-aware ASP, the landscape properties of problem instances are associated with the performance of optimization algorithms.
For this, the most common way is by characterizing the landscape characteristics or high-level properties of a problem instance, such as its global structure, multi-modality and separability~\cite{mersmann2010benchmarking}.
Nonetheless, an accurate characterization of these high-level properties is challenging without expert knowledge.
To facilitate the landscape characterization, ELA has been introduced to capture the low-level properties of a problem instance, e.g., $y$-distribution, level set and meta-model~\cite{mersmann2011exploratory}.
It has been shown that these ELA features are sufficiently expressive in accurately classifying the BBOB functions according to their corresponding problem classes~\cite{renau2021towards} and also informative for algorithm selection purposes~\cite{munoz2015algorithm, kerschke2019automated}.

In the ELA, landscape features are computed primarily using a design of experiment (DoE) of some sample points $\mathcal{X}=\{x_1,...,x_n\}$ evaluated on an objective function $h$, i.e., $h \colon \mathbb{R}^d \rightarrow \mathbb{R}$, with $x_i \in  \mathbb{R}^d$, $n$ represents sample size, and $d$ represents function dimensionality.
In this work, we compute the ELA features using the \texttt{pflacco} package~\cite{prager2022pflacco}, which was developed based on the \texttt{flacco} package~\cite{kerschke2019flacco}.
With more than 300 ELA features that can be computed, we consider only the ELA features that can be cheaply computed without additional re-sampling, similar to the work in~\cite{long2022learning}, and disregard the ELA features that only concern the DoE samples (altogether four of the principal component analysis features).

While we are fully aware that the ELA features are highly sensitive to sample size~\cite{munoz2022analyzing} and sampling strategy~\cite{skvorc2021effect}, these aspects are beyond the scope of this research.
Throughout this work, we consider the Sobol' sampling technique~\cite{sobol1967distribution} based on the results in~\cite{renau2020exploratory}.

\subsection{Instance Space Analysis}
In general, ISA is a methodology of benchmarking algorithms and assessing their strengths and weaknesses based on clusters of problem instances~\cite{mario_andres_munoz_2023_7700434}.
The instance space refers to the set of all possible problem instances that can be used to evaluate the performance of such an optimization algorithm. 
The fundamental idea behind ISA is to model the relationship between the structural properties of a given problem instance and the performance of a set of algorithms.
Through this approach, footprints can be constructed for each algorithm, which are essentially regions in the instance space where statistically significant performance improvements can be inferred.

In our proposed approach, a similar concept is utilized, namely to find such problem instances that resemble specific expensive real-world problems, to allow the comparison and benchmarking of various algorithms on a highly specific instance space. 
This differs from other related works, such as the Melbourne algorithm test instance library with data analytics (MATILDA) software~\cite{smith2023instance}, because while other works try to create a space-covering of instances for a general benchmarking purpose and comparison of algorithms, we aim to generate highly specialised benchmarks sets that are cheap-to-evaluate and representative for expensive real-world black box problems.
Generating such domain-specific benchmarks would allow us to search and optimize specific algorithm configurations that are better applicable to a specific problem domain. 
Furthermore, it also allows a better understanding of the expensive black-box problem through the analysis of different optimization landscapes that represent the found instance space.

\subsection{Genetic Programming}

Principally, GP can be considered as a search heuristic for computer program synthesis that is inspired by neo-Darwinian evolution.
The first work on GP was done by Forsyth~\cite{doi:10.1108/eb005587}, Cramer~\cite{cramer1985representation} and Hicklin~\cite{Hicklin}. 
Later, work by Koza~\cite{Koza89, koza1990genetic, Koza:1992:GPP:138936, koza1994genetic} significantly popularized the field of GP. 
As proposed by Koza, GP is traditionally used with trees as program representation.
A typical application of GP is to solve optimization problems that can be formulated as:
$\underset{\mathbf{t}}{\mathrm{\arg\min}} \enskip f(\mathbf{t}), \mathbf{t} \in \mathcal{T}$,
where $\mathbf{t} = (t_1, \cdots, t_n)$ represents a decision vector (also known as individual or solution candidate) in evolutionary algorithms (EAs).
Similar to other EAs, GP evolves a population of solution candidates by following the principle of the survival of the fittest and utilizing biologically-inspired operators. 
The feature distinguishing GP from other EAs is the variable-length representation for $\mathbf{t}$, instead of a fixed-length representation.

As one of the powerful optimization techniques, GP has been widely used to solve regression problems by searching through a space of mathematical expressions. 
In fact, GP-based symbolic regression (SR)~\cite{Tackett1995} is popular as an interpretable alternative to black-box regression methods, where GP is used to search for an explicit mathematical expression for a given dataset.
By producing a mathematical expression that can be easily understood by humans, SR has proven to be a valuable tool in engineering applications, where it is important to comprehend the relationship between different decision variables.

In fact, GP-based SR has been widely used as a surrogate model to replace expensive simulation problems in engineering. 
Using GP-based SR, for instance, Singh et al.~\cite{gp/app/singh2007genetic} quantified the non-linear relationship among multiple decision variables in the rotating beams problems, Jalal et al.~\cite{gp/app/Jalal2013} modeled the strength enhancement of concrete cylinders, and Kalita et al.~\cite{gp/app/Kalita2020} created laminated composites model for multi-objective multi-fidelity problems. 
Recently, Yang and Affenzeller~\cite{yang2023emo} utilized GP-based SR and its variants to build surrogate models for multi-objective optimization.  

While GP-based SR was mainly used as a surrogate model to either quantify the relationship between different decision variables or replace expensive optimization problems in previous work, we focus on utilizing canonical GP~\cite{tian2020arecommender} to create functions with specific target ELA features in this work.

\subsection{Generating Black-Box Optimization Problems}\label{sec:rfg}

Apart from the expertly designed benchmarking test suites, Tian et al. introduced a SR approach in generating continuous black-box optimization problems \cite{tian2020arecommender}.
In their work, a function generator was proposed to generate problem instances of different complexity in the form of tree representations that serve as training samples for a recommendation model.
More specifically, the function generator constructs a tree representation by randomly selecting mathematical operands and operators from a predefined pool, where each operand and operator has a specific probability of being selected.
In this way, any arbitrary number of functions can be quickly generated.
To improve the functional complexity that can be generated, such as noise, multi-modal landscape and complex linkage between variables, a \textit{difficulty injection} operation was included to modify the tree representation.
Furthermore, a \textit{tree-cleaning} operation was considered to simplify the tree representation by eliminating redundant operators.
In the remainder of this paper, we refer to this function generator as random function generator (RFG).

In fact, functions generated by the RFG indeed have landscape characteristics different from the BBOB test suite and complement the coverage of BBOB functions in the instance space~\cite{skvorc2021acomplementarity}.
In the work by Long at el., the landscape characteristics of functions generated by the RFG were compared to several real-world automotive crashworthiness optimization problem instances~\cite{long2022learning}.
It has been shown that, as far as landscape characteristics are concerned, some functions generated by the RFG belong to the same problem class as the automotive crash problems.

In addition to GP and RFG approach, a recent paper proposed to make use of affine combinations of BBOB functions and showed that these new functions can help fill empty spots in the instance space~\cite{dietrich2022increasing}.
Essentially, a new function is constructed via a convex combination of two selected BBOB functions, using a weighting factor to control the interpolation.
Extensions of this work have generalized the approach to affine combinations of more functions (not limited to only two functions) and shown their potential for the analysis of automated algorithm selection methods~\cite{affine_bbob_gecco, many_affine_bbob}.

\section{Methodology}\label{methodology}

In brief, we develop our GP-based function generator (we simply refer this as GP in the remainder of this paper) based on the RFG and canonical GP approach.
Precisely, we consider the mathematical operands and operators similar to those used in the RFG with slight modifications, as summarized in Table~\ref{tab:gp_x_space}.
Following this, the GP search space consists of the terminal space $\mathcal{S}$ (operands) and function space $\mathcal{F}$ (operators), i.e., $\mathcal{T} = \{\mathcal{S}  \cup \mathcal{F} \}$.
Unlike typical GP-based SR method, where each design variable $x_i$ is separately treated (as terminal), we consider tree-based math expression, that is, a vector-based input $\mathbf{t} = (t_1, \cdots, t_d)$, to facilitate a comparison with the RFG.

\begin{algorithm2e}[]
\KwIn{
objective function $f(.)$,
population size $n_{pop}$,
crossover rate $p_c$,
mutation rate $p_m$,
maximum number of generation $g_{\mathrm{max}}$
;}
\KwOut{
optimal solution $t^*$
}
$g \longleftarrow 0$\;
$Pop(g) \longleftarrow $ InitializePopulation ($n_{pop}$) \;
Evaluate($Pop(g), f $)\;
\While{$g \leq g_{\mathrm{max}}$}{
    $Pop'(g) \longleftarrow $ MatingSelection($Pop(g)$) \;
    $Pop''(g) \longleftarrow $ Variation($Pop'(g)$, $p_c$, $p_m$) \;
    Evaluate($Pop''(g), f $) \;
    $Pop(g+1) \longleftarrow Pop(g)$ \;
    $g \longleftarrow g + 1$\;
}
$t^*\longleftarrow  \arg\min f(t)$ where $t \in Pop(g)$\;
\caption{{\bf Pseudocode of canonical GP. \emph{Variation} operation includes crossover/recombination and mutation.} \label{alg:gp}}
\end{algorithm2e}

\begin{table*}[!t]
  \centering
  \caption{List of notations and their meaning, syntax, protection rules (if any) and probability of being selected for the GP sampling (only during the first generation).}\label{tab:gp_x_space}
    \begin{tabular}{ccccc}
    \toprule
    Notation & Meaning & Syntax & Remark/Protection & Probability\\
    \midrule
    \multicolumn{5}{c}{Operands ($\mathcal{S}$)} \\
    \midrule
    \texttt{x} & Decision vector & $(x_1, \dots ,x_d)$ &       & 0.6250 \\
    \texttt{a} & A real constant & $a$   &  $\mathcal{U}(1,10)$     & 0.3125 \\
    \texttt{rand} & A random number & $rand$ &  $\mathcal{U}(1,1.1)$     & 0.0625 \\
    \midrule
    \multicolumn{5}{c}{Operators ($\mathcal{F}$)} \\
    \midrule
    \texttt{add} & Addition & $a+x$ &       & 0.1655 \\
    \texttt{sub} & Subtraction & $a-x$ &       & 0.1655 \\
    \texttt{mul} & Multiplication & $a \cdot x$ &       & 0.1098 \\
    \texttt{div} & Division & $a/x$ & Return $1$, if $\left | x \right |\leq10^{-20}$ & 0.1098 \\
    \texttt{neg} & Negative & $-x$  &       & 0.0219 \\
    \texttt{rec} & Reciproval & $1/x$ & Return $1$, if $\left | x \right |\leq10^{-20}$ & 0.0219 \\
    \texttt{multen} & Multiplying by ten & $10x$ &       & 0.0219 \\
    \texttt{square} & Square & $x^2$ &       & 0.0549 \\
    \texttt{sqrt} & Square root & $\sqrt{\left | x \right |}$ &       & 0.0549 \\
    \texttt{abs} & Absolute value & $\left | x \right |$ &       & 0.0219 \\
    \texttt{exp} & Exponent & $e^x$ &       & 0.0219 \\
    \texttt{log} & Logarithm & $\ln \left | x \right |$ & Return $1$, if $\left | x \right |\leq10^{-20}$ & 0.0329 \\
    \texttt{sin} & Sine  & $sin(2\pi x)$ &       & 0.0329 \\
    \texttt{cos} & Cosine & $cos(2\pi x)$ &       & 0.0329 \\
    \texttt{round} & Rounded value & $\left \lceil x \right \rceil$ &       & 0.0329 \\
    \texttt{sum} & Sum of vector & $\sum_{i=1}^{d}x_i$ &       & 0.0329 \\
    \texttt{mean} & Mean of vector & $\frac{1}{d}\sum_{i=1}^{d}x_i$ &       & 0.0329 \\
    \texttt{cum} & Cumulative sum of vector & $(\sum_{i=1}^{1}x_i,\dots,\sum_{i=1}^{d}x_i)$ &       & 0.0109 \\
    \texttt{prod} & Product of vector & $\prod_{i=1}^{d}x_i$ &       & 0.0109 \\
    \texttt{max} & Maximum value of vector & $\textup{max}_{i=1,\dots,d}x_i$ &       & 0.0109 \\
    \bottomrule
    \end{tabular}%
\end{table*}%

Regarding the GP aspect, we consider the canonical GP as shown in Algorithm \ref{alg:gp} and the distributed evolutionary algorithms in Python (DEAP) package~\cite{DEAP_JMLR2012}.
The descriptions of our GP function generator are as follows.

\paragraph{Input}
A set DoE samples $\mathcal{X}$ and the ELA features of the target functions are used as input for the GP pipeline. 

\paragraph{Objective function}
In the GP-system, our optimization target is to minimize the differences between the ELA features of an individual and the target function.
Before the ELA feature computation, we normalize the objective values (by min-max scaling) to remove inherent bias as proposed in~\cite{prager2023nullifying}.
Furthermore, we normalize the ELA features (by min-max scaling ) before the distance computation, to ensure that all ELA features are within a similar scale range.
For this, we consider the minimum and maximum values from a set of BBOB functions ($24$ BBOB functions, $5$ instances each).
Based on the same set of BBOB functions, we identify and filter out ELA features that are highly correlated in a similar fashion to the work in~\cite{long2022learning}. 
And we consider equal weighting of ELA features in this work.

\paragraph{Infeasible Solutions}
An individual is considered to be invalid, if any of the four following conditions is fulfilled:
\begin{enumerate}
    \item Error when converting the tree representation to an executable Python expression,
    \item Bad objective values, e.g., infinity, missing or single constant value,
    \item Error in ELA computation, e.g., due to equal fitness in all samples, and
    \item Invalid distance, caused by missing value in ELA feature.
\end{enumerate}
All invalid trees are penalized with a large fitness of $10\,000$.

\paragraph{Initialize Population}
In the first generation, we initialize the initial population using random sampling. We make use of a population size of $50$ for computational reasons. The tree depth is limited between $3$ and $12$.
Similar to the RFG, we assigned each operand and operator a probability of being selected (Table~\ref{tab:gp_x_space}).
This step is crucial in reducing the number of invalid trees (e.g., errors 1), as we would like to keep the first population as healthy as possible.
Furthermore, whenever an invalid tree is generated, we will do re-sampling, meaning that this invalid tree will be replaced by generating a new tree.
As such, we ensure that the initial population is free from invalid trees (only errors 1 and 2).

\paragraph{Mating Selection and Variation}
We consider the tournament selection with tournaments of size $5$, subtree crossover with a crossover probability of $0.5$, and subtree mutation with a mutation probability of $0.1$.
Other remaining hyperparameters are set to default settings.
While hyperparameter tuning could potentially improve our results further, we decide to leave it for future work.

\paragraph{Output}
A set of individuals generated by during the GP runs and their corresponding fitness values.

\section{Experimental Setup}\label{experiment}

In this work, we tested our pipeline taking as target functions all the $24$ BBOB functions (one by one) of three different dimensionalities $d=2$, $5$ and $10$ (or simply $2d$, $5d$ and $10d$).
We consider a DoE size of $150d$ samples and the search domain $[-5,5]^d$.

To reliably capture the landscape characteristics, we compute ELA features in a bootstrapping manner (using only $80\%$ of the DoE samples and $5$ repetitions with different random seeds).
As for the optimization objective or individual fitness in the GP system, we consider minimizing the average Wasserstein distance between the ELA features of the first five BBOB instances and the evaluated individual (5 bootstrapped samples for each feature, for each instance).
Due to computational limits, we perform only one run for each target function in each dimensionality.

\paragraph{Reproducibility} To ensure the reproducibility of our work, the codes used to generate, process and visualize the experiments have been made available in a Zenodo repository~\cite{zenodo_repo}. 
This repository also contains the raw data and more figures which could not be included in this paper due to space limitations.

\section{Results}\label{results}

\subsection{Performance of GP}

\begin{figure}[!ht]
    \centering
    \includegraphics[width=0.48\textwidth,trim=10mm 11mm 5mm 9mm,clip]{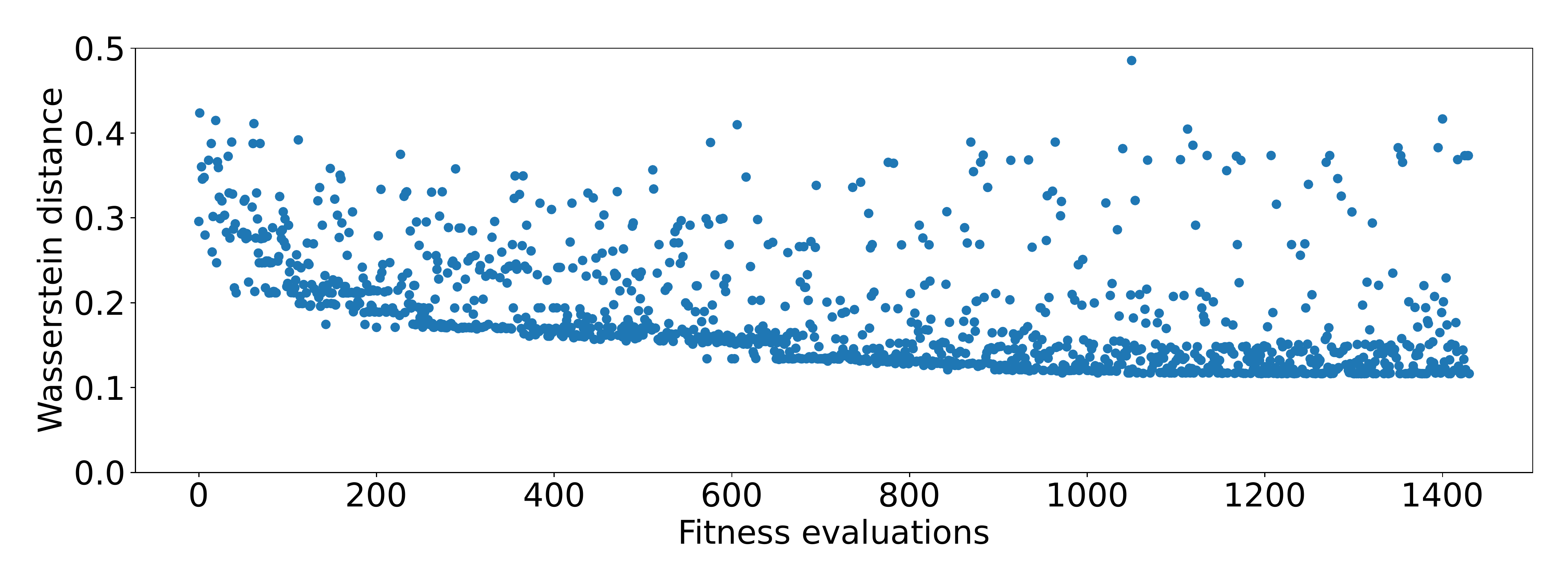}
    \caption{GP convergence for target F1 (sphere), in $2d$.}
    \label{fig:convergence_f1_2d}
\end{figure}

We start by analysing the functions generated during the GP runs with a $2d$ BBOB function as their target. 
In Figure~\ref{fig:convergence_f1_2d}, we show the convergence trajectory of a single run on F1.
From this figure, we clearly see that GP manages to improve over the initial population (first 50 evaluations), as time goes on. 
Note that, while it has a budget of 50 generations of 50 individuals each, the combination of a crossover rate of $0.5$ with a mutation rate of $0.1$ means that there is a probability of $0.45$ that a selected individual is not modified in any way, and just copied to the next generation without being evaluated. 
Additionally, Figure~\ref{fig:convergence_f1_2d} does not include the infeasible solutions, which make up $2\%$ of all evaluations in this run. 

\begin{figure*}[!ht]
    \centering
    \includegraphics[width=\textwidth,trim=10mm 11mm 5mm 9mm,clip]{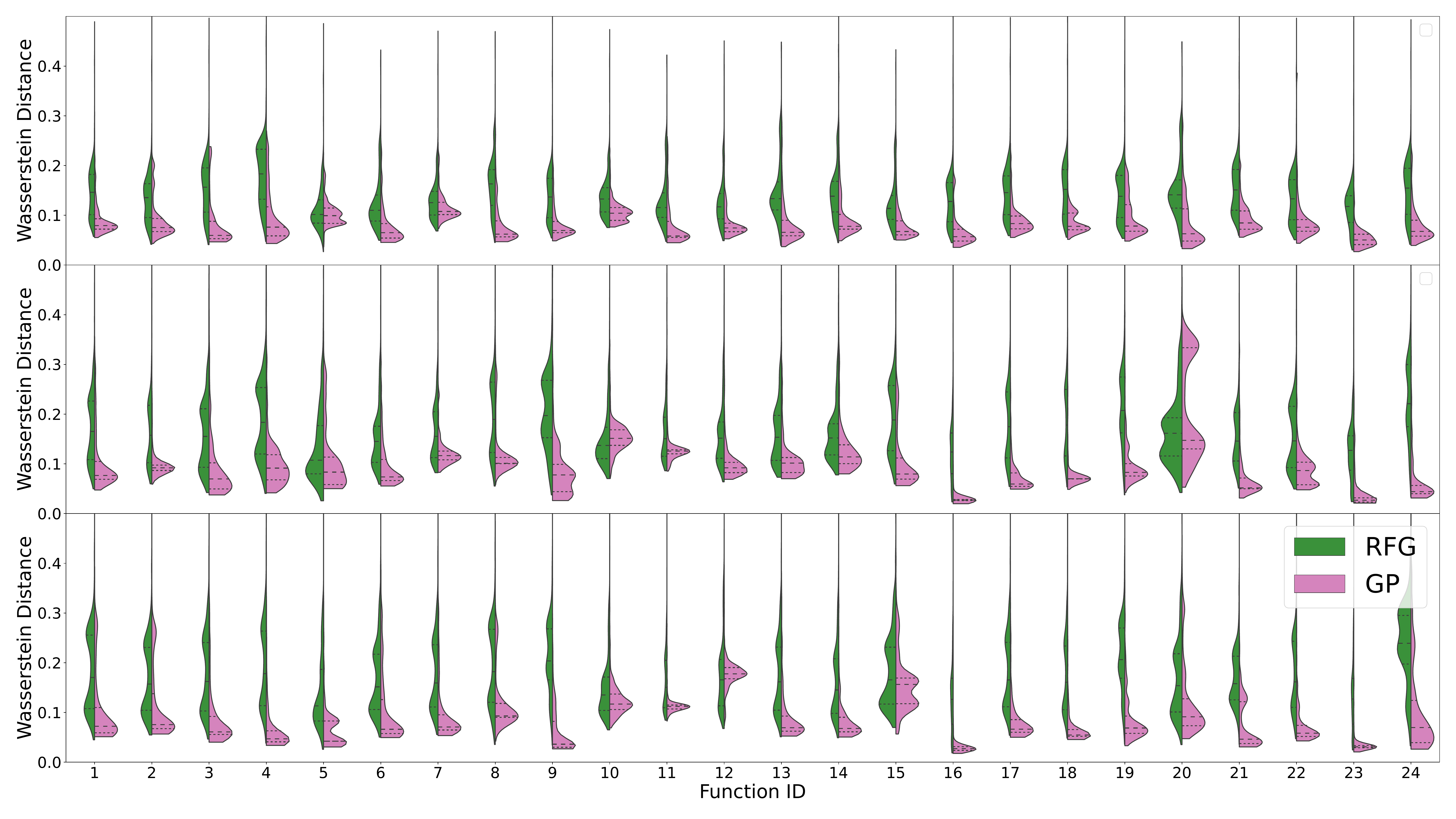}
    \caption{Distribution of fitness values (Wasserstein distance) of the set of functions from the RFG and the GP 
    runs with the specified BBOB target functions (horizontal axis). Rows correspond to dimensionalities: $2d$, $5d$ and $10d$ (from top to bottom).}
    \label{fig:gp_vs_rfg}
\end{figure*}

To give some context to the fitness values shown in Figure~\ref{fig:convergence_f1_2d}, we compare functions generated via our GP approach with functions generated by the RFG (see Section~\ref{sec:rfg}). 
For this purpose, we generate $1\,000$ valid functions (using the RFG) for each dimensionality and measure their Wasserstein distance to each of the $24$ target BBOB functions. 
Then, we compare these distances to those of the functions generated during our GP runs ($\sim1\,400$ functions) to the corresponding target problem. 
This is visualized in Figure~\ref{fig:gp_vs_rfg}, from which we see that the lower end (i.e., generated functions with low distance to target functions) of the GP distribution is almost always better than that of the RFG, with some exceptions, e.g., F12 (Bent Cigar) in $10d$.

\begin{figure*}[!ht]
    \centering
    \includegraphics[width=0.90\textwidth]{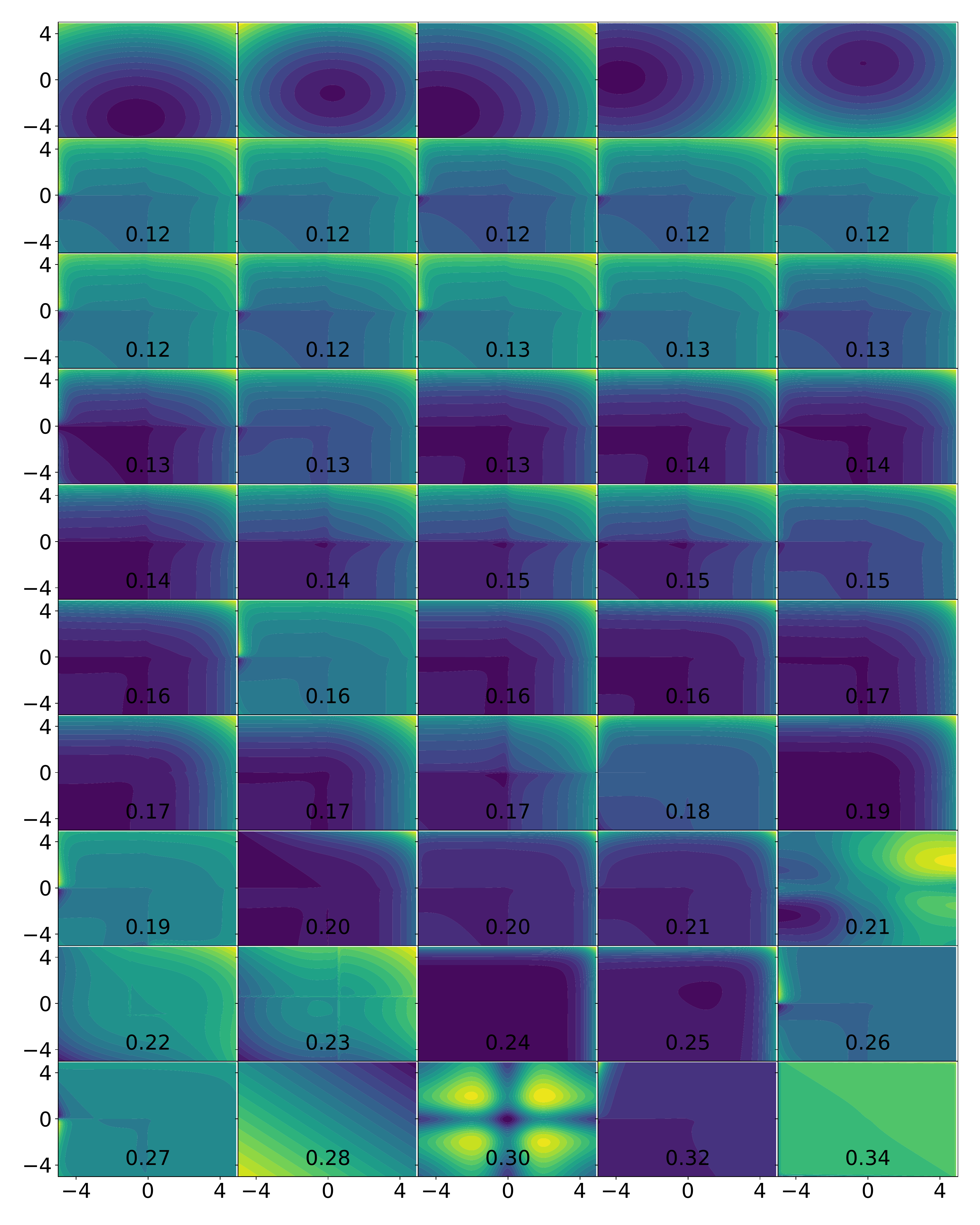}
    \caption{Grid of functions generated by the GP procedure with target function F1 (sphere) in $2d$, of which 5 instances are plotted in the first row. The rows below are GP-generated problems selected by ranking their Wasserstein distance to the target feature vector (indicated by the value in each subfigure) and taking a linear spacing in this ranking from the best (top left) to the worst (bottom right).}
    \label{fig:F1_grid}
\end{figure*}

For these GP runs, we can also visualize the resulting function landscapes to identify how much they resemble the target BBOB function. 
This is shown in Figure~\ref{fig:F1_grid} for F1 (sphere) in $2d$, where the 5 BBOB instances are plotted in the first row, followed by 45 GP-generated functions. 
These functions are selected by sorting their fitness values and taking a linear spacing in the rank values between the best and worst functions, to show a range of generated functions of varying quality. 
From visual inspection, we notice that even the best functions (with the smallest Wasserstein distance) do not quite represent a sphere as we might have expected.
  
\begin{figure*}[!ht]
    \centering
    \includegraphics[width=0.90\textwidth]{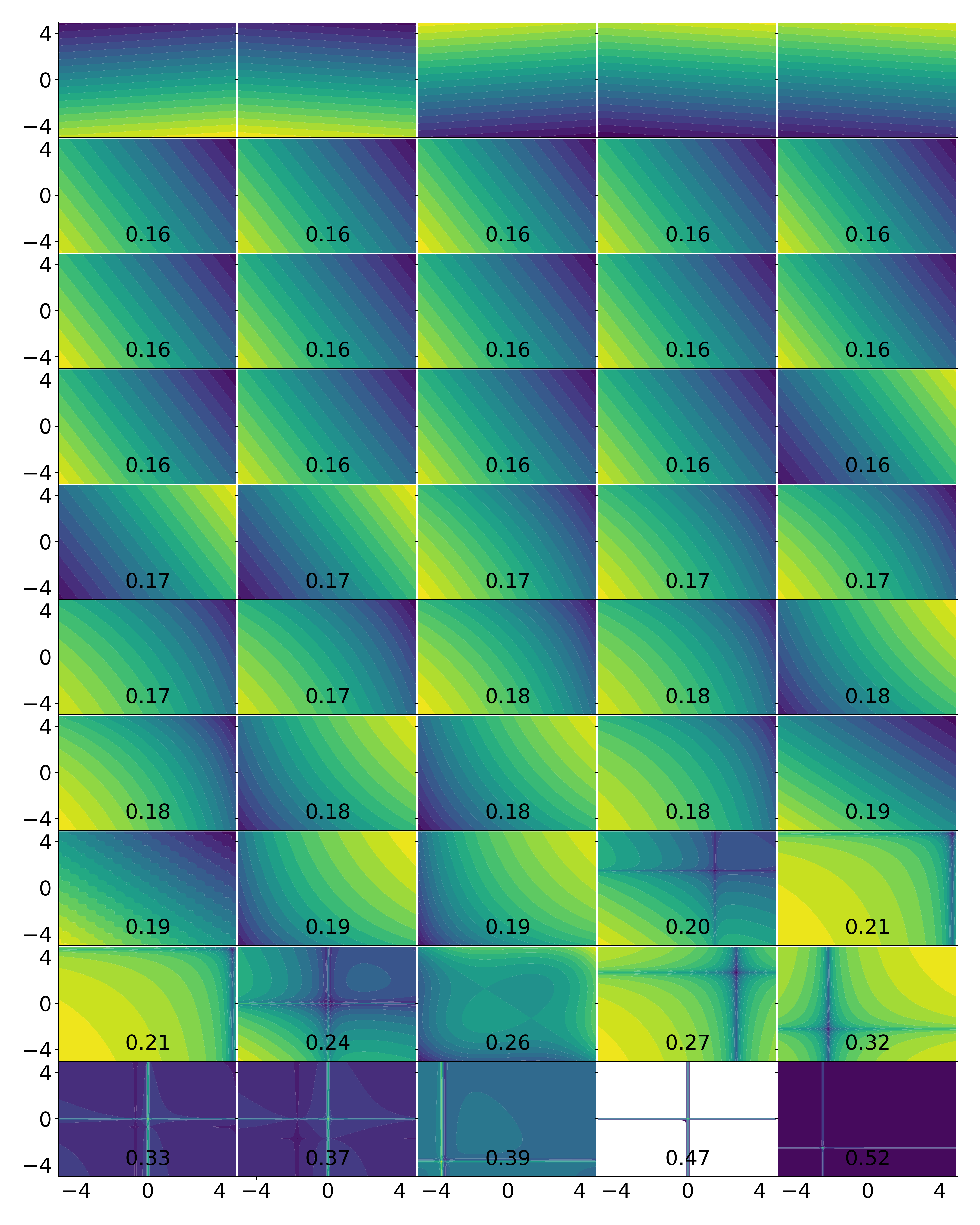}
    \caption{Grid of functions generated by the GP procedure with target function F5 (linear slope) in $2d$, of which 5 instances are plotted in the first row. The rows below are GP-generated problems selected by ranking their Wasserstein distance to the target feature vector (indicated by the value in each subfigure) and taking a linear spacing in this ranking from the best (top left) to the worst (bottom right).}
    \label{fig:F5_grid}
\end{figure*}

In a similar fashion, we can create the same visualization for other functions, e.g., F5 (linear slope), as shown in Figure~\ref{fig:F5_grid}, where we observe a much closer matching between the target and generated functions.
It is, however, interesting to note that some generated functions, e.g., row 7 column 5 (which represents function $\frac{x_0+x_1}{2}$), appear visually similar to the target, but nonetheless have a relatively large fitness value. 
\textit{This raises the question of whether the distance to the target distribution in ELA space really captures the intuitive global properties of the linear slope problem}.

\subsection{Investigating the ELA Space}

\begin{figure*}[!ht]
    \centering
    \includegraphics[width=\textwidth,trim=12mm 11mm 9mm 8mm,clip]{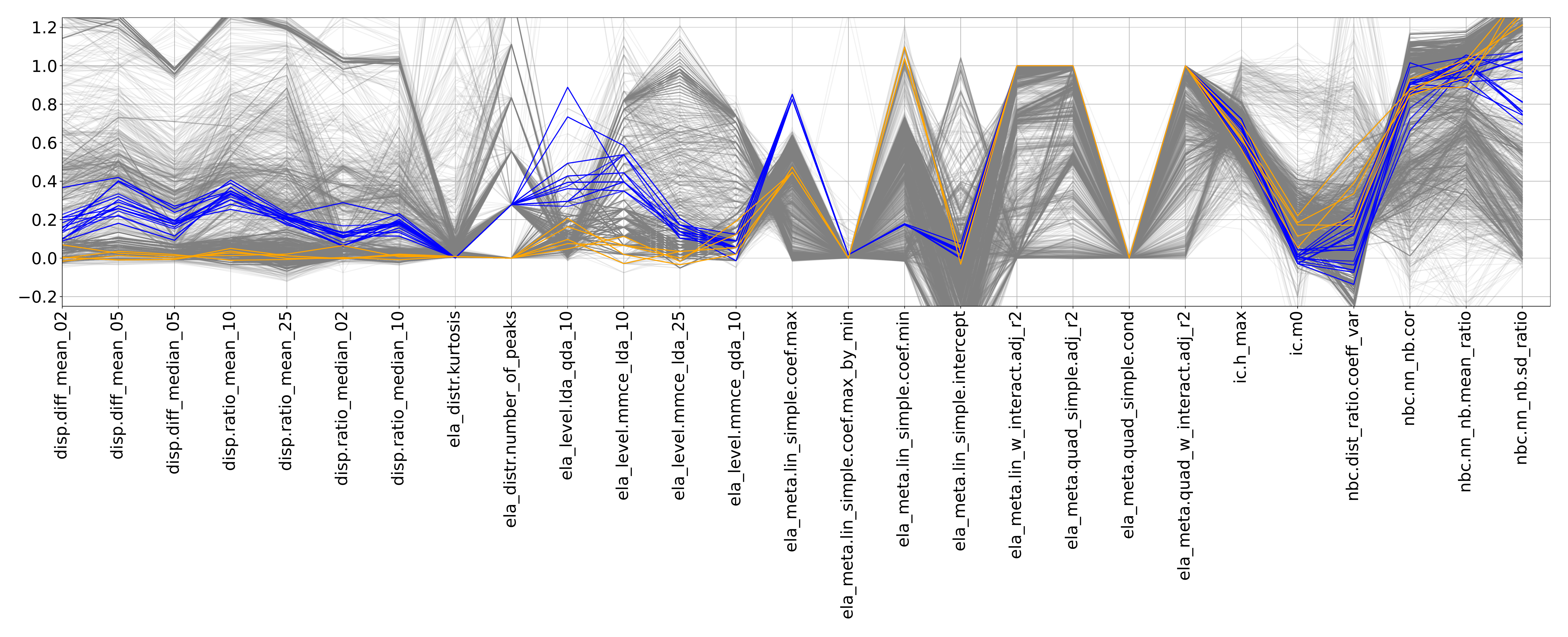}
    \caption{Parallel coordinate plot of the ELA features (one line for each bootstrapped DoE) for the GP-generated functions and target function F5 (linear slope) in $2d$, of which 5 instances are plotted in blue. The orange lines highlight an example of GP-generated function, corresponding to $\frac{x_0+x_1}{2}$, which has a Wasserstein distance of $0.19$ to the target. }
    \label{fig:F5_parallel}
\end{figure*}

To understand why the distance between this linear slope and the target function is relatively large, we need to look at the individual ELA features. 
This can be done through a parallel coordinate plot, as shown in Figure~\ref{fig:F5_parallel}. 
From this figure, we can see that mostly the \textit{ela\_meta.lin\_simple.coef.min} and \textit{ela\_meta.lin\_simple.coef.max} are different between the target function and generated function $\frac{x_0+x_1}{2}$, indicating that the steepness of the function might be different.
However, for the linear slope function, this should not have a large impact as the global properties are mostly preserved. 
This shows that different ELA features are crucial for different types of target functions. 
One direction to mitigate such problems might be to analyse the spread of ELA feature values over a large set of instances for each BBOB function (see discussion around Figure~\ref{fig:std_bbob} below).

To better understand the similarities between functions, previous work~\cite{long2023bbob, skvorc2021acomplementarity, renau2021towards} often makes use of a dimensionality reduction approach to visualize the high-dimensional ELA space. 
By utilizing the uniform manifold approximation mapping (UMAP)~\cite{umap_mcinnes2018} method, we visualize the ELA space filled by the newly GP-generated functions relative to the existing BBOB problems.

\begin{figure}[!ht]
    \centering
    \includegraphics[width=0.48\textwidth,trim=10mm 11mm 5mm 9mm,clip]{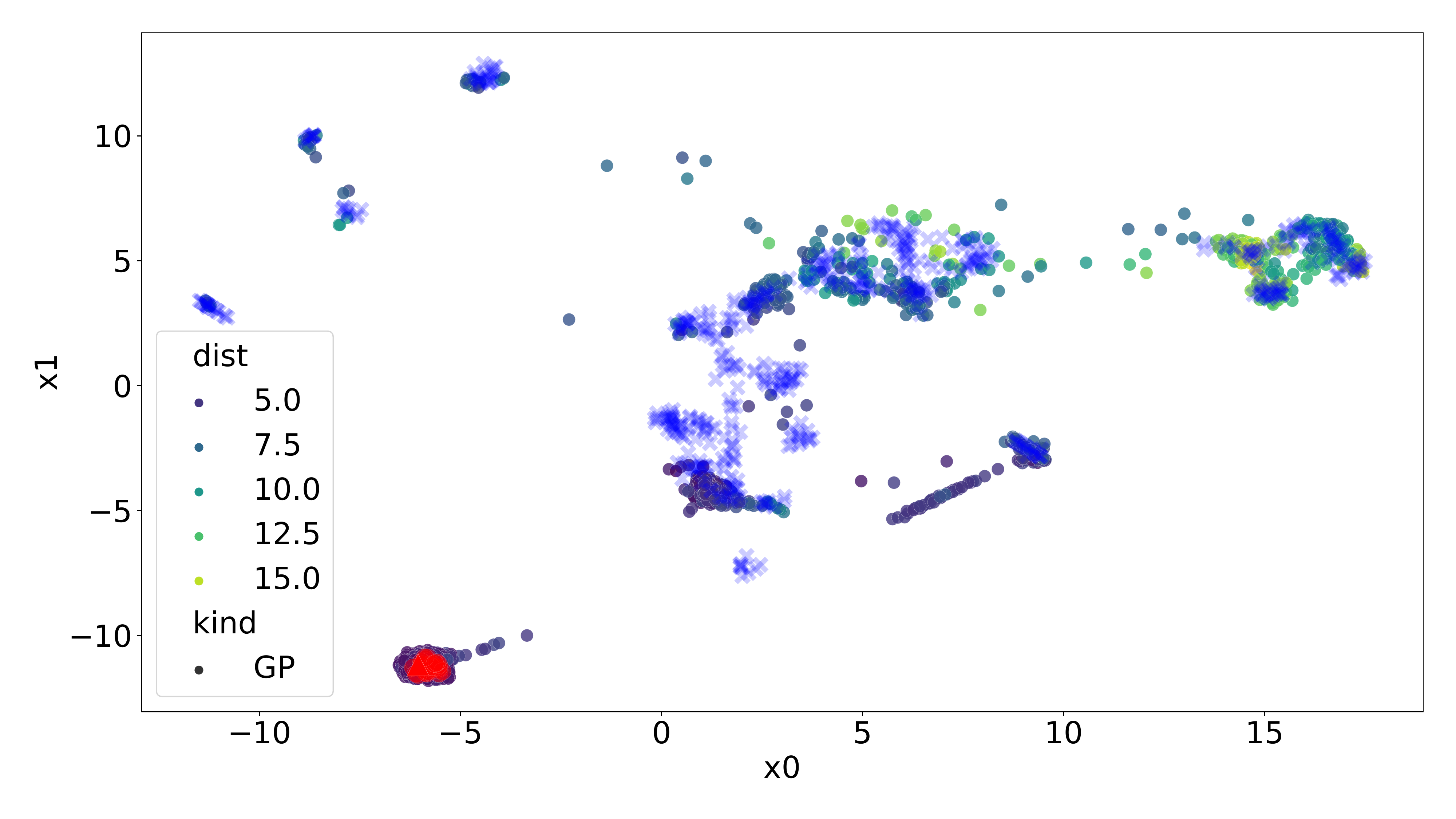}
    \caption{UMAP projection of the GP-generated functions with the target function F5 (linear slope) in $2d$ ELA space. 
    The mapping is based on BBOB instances only, which are highlighted in blue crosses. The target F5 is highlighted in red, with the mean feature vector across the five instances indicated as a red triangle. The dots correspond to the GP-generated problems, where the colour is the cityblock or Manhattan distance to the target vector (here, coloring based on Wasserstein distance is challenging).}
    \label{fig:umap_f5}
\end{figure}

To achieve this, we first create the mapping using only the feature representations from the $24$ BBOB problems (all five bootstrapped repetitions on each instance).
Next, we apply this fixed map on the GP-generated functions (from one run of the GP). 
The resulting plot for the target function F5 (linear slope) is shown in Figure~\ref{fig:umap_f5}, where we see that most GP-generated functions are indeed clustered together around the target. 

\subsection{Distances in Feature Space}

Our suspicion for why the distances in ELA space do not directly seem to correlate to our visual understanding of high-level properties might be that all ELA features are weighted equally. 
This means that even features, which are more sensitive to small deviations, can have the same impact as features that might be considered crucial to characterize a function e.g., a linear slope.

\begin{figure}[!ht]
    \centering
    \includegraphics[width=0.48\textwidth,trim=9mm 10mm 35mm 9mm,clip]{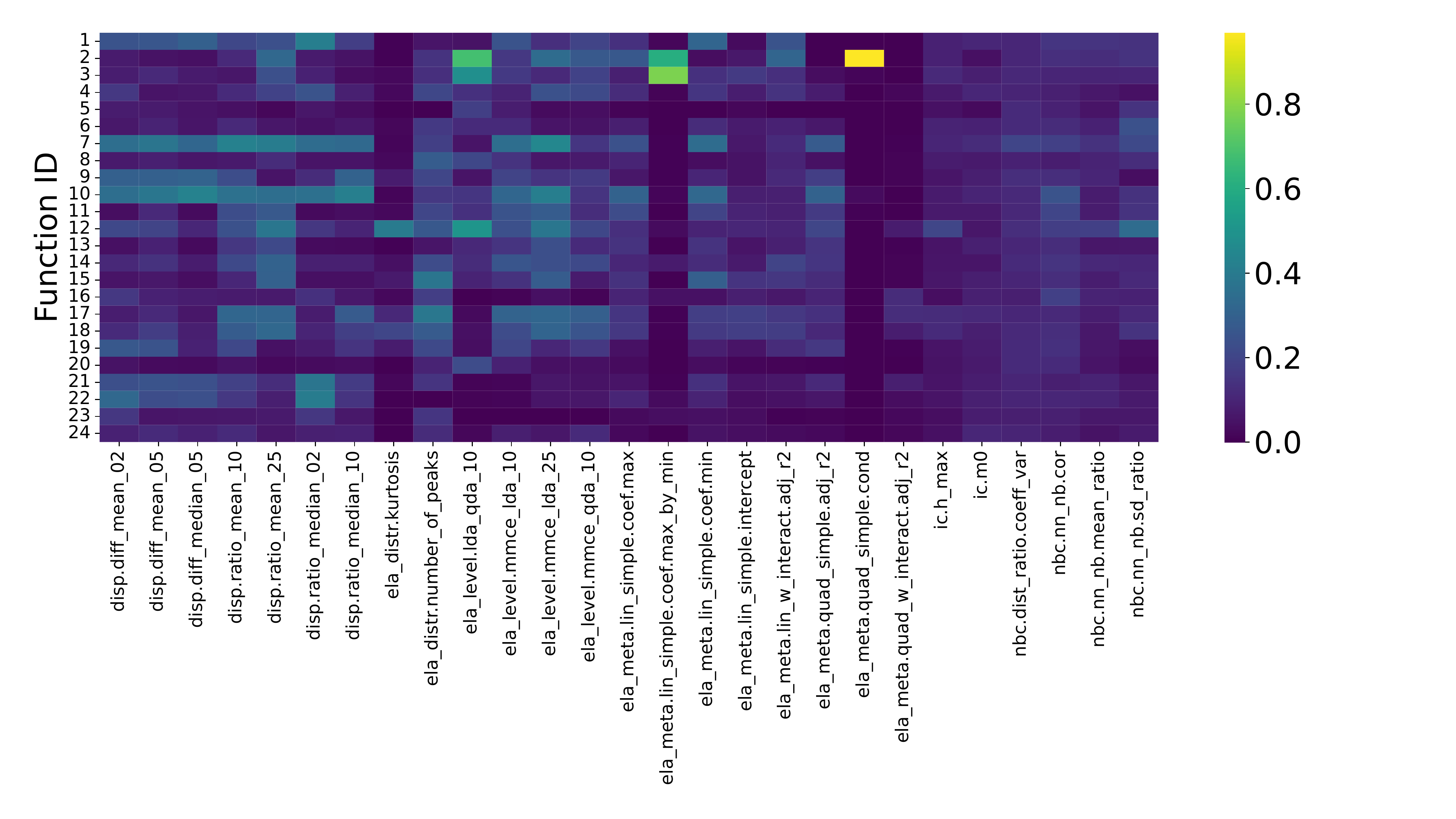}
    \caption{Relative standard deviation (of the normalized values) of each ELA feature for each BBOB function. Lighter color represents a larger deviation.}
    \label{fig:std_bbob}
\end{figure}

To gain insight into \textit{which features might be more important for a given function}, we analyze the relative standard deviation of each ELA feature within instances of the same BBOB function and visualized the results in Figure~\ref{fig:std_bbob}.

While Figure~\ref{fig:std_bbob} shows us the variance of each ELA feature within the target functions, it is important to relate this to the deviations observed in the GP-generated functions.  
Following this, we look at the average standard deviation of each ELA feature across all functions sampled during a single GP run and compute the absolute difference to the values seen on the corresponding BBOB function target. 

\begin{figure}[!ht]
    \centering
    \includegraphics[width=0.48\textwidth,trim=9mm 10mm 35mm 9mm,clip]{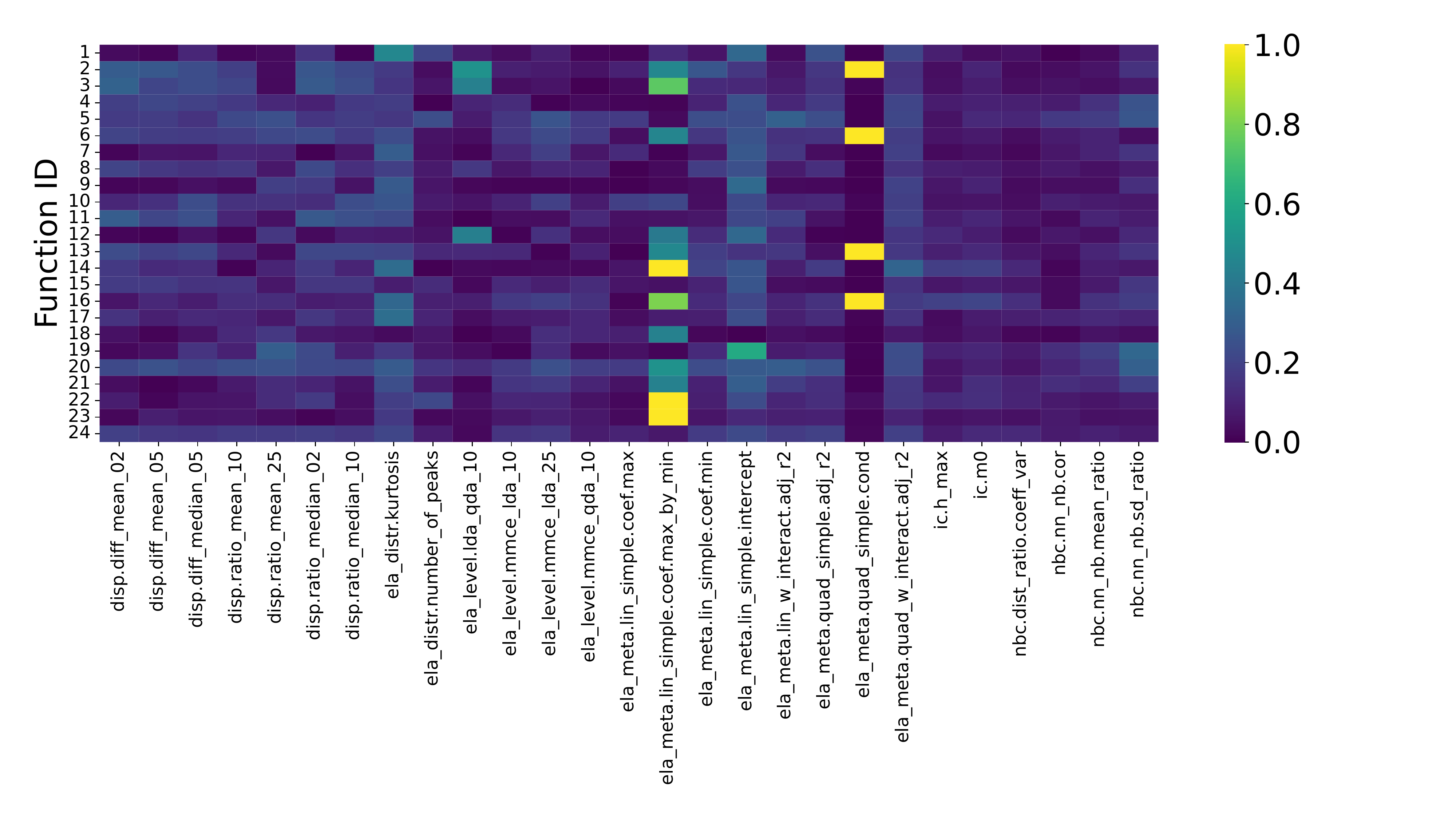}
    \caption{Absolute difference in relative standard deviation (of the normalized values) of each ELA feature for each BBOB function to the functions generated by the GP run with the respective target. Lighter color represents a larger deviation.}
    \label{fig:std_bbob_gp}
\end{figure}

The resulting differences are shown in Figure~\ref{fig:std_bbob_gp}, where we can see that some features, such as the max-by-min of the linear coefficient (\textit{ela\_meta.lin\_simple.coef.max\_by\_min}), are significantly more variable in the GP-generated functions than in BBOB.
This indicates that these ELA features might be very important for the distance measurement. 

\begin{figure}[!ht]
    \centering
    \includegraphics[width=0.4\textwidth,trim=9mm 10mm 20mm 9mm,clip]{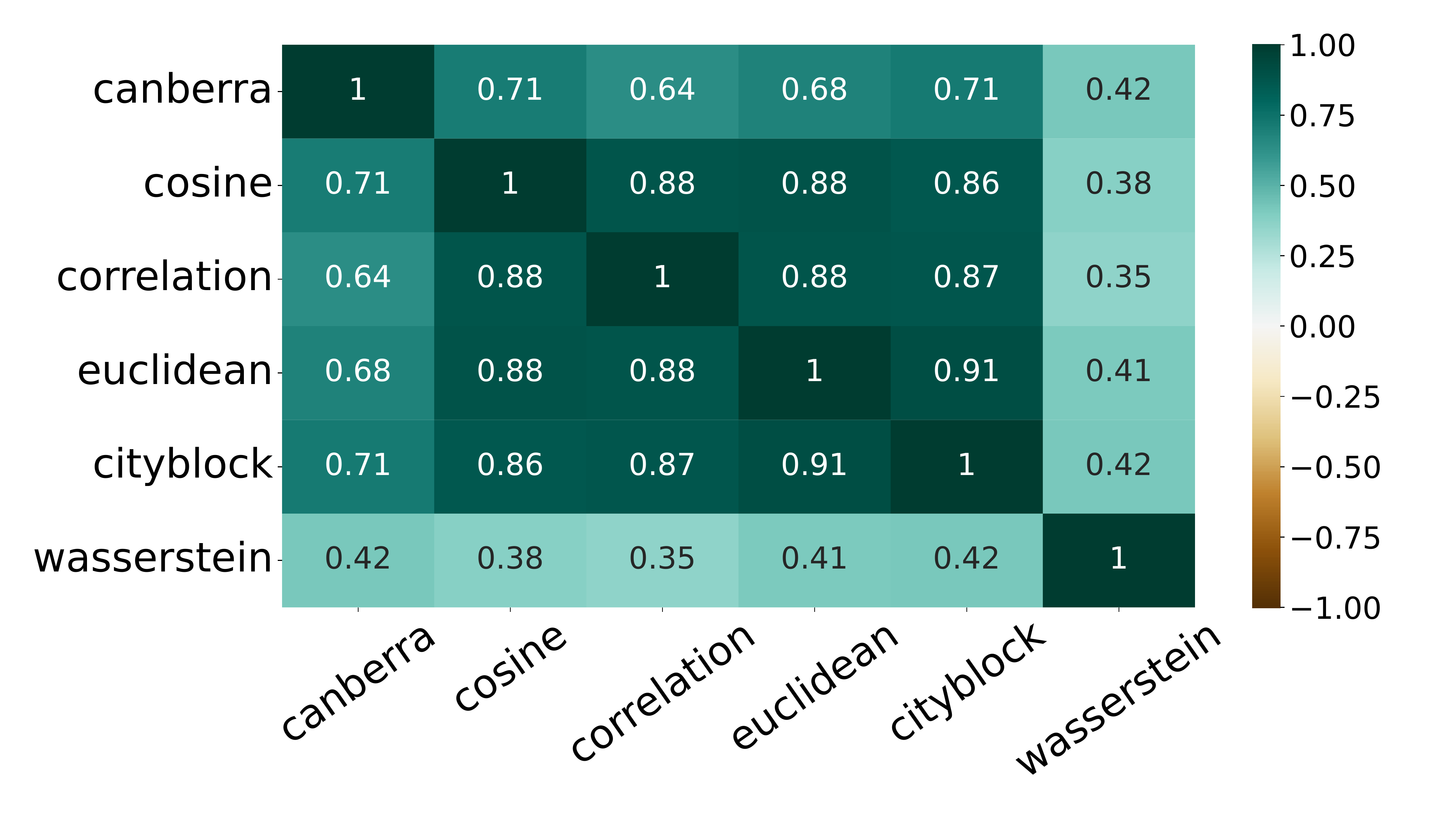}
    \caption{Kendall-tau correlation between six distance measures on the BBOB instances. }
    \label{fig:dist_corr}
\end{figure}

In our GP experiments, we consider the average Wasserstein distance to determine the fitness value. 
This choice was made, since each ELA feature can be considered as a random variable, for which a statistical distance measure would be appropriate. 
Alternatively, we could make use of a regular distance measure, based on the mean values of these ELA feature distributions.
To determine the impact the choice of distance measure might have, we consider the Kendall-tau correlation between a set of six metrics on the pairwise distances between the BBOB instances, consisting of the Canberra, cosine, correlation, Euclidean, cityblock and Wasserstein distance, as visualized in Figure~\ref{fig:dist_corr}. 

\begin{figure}[!ht]
    \centering
    \includegraphics[width=0.48\textwidth,trim=10mm 11mm 5mm 9mm,clip]{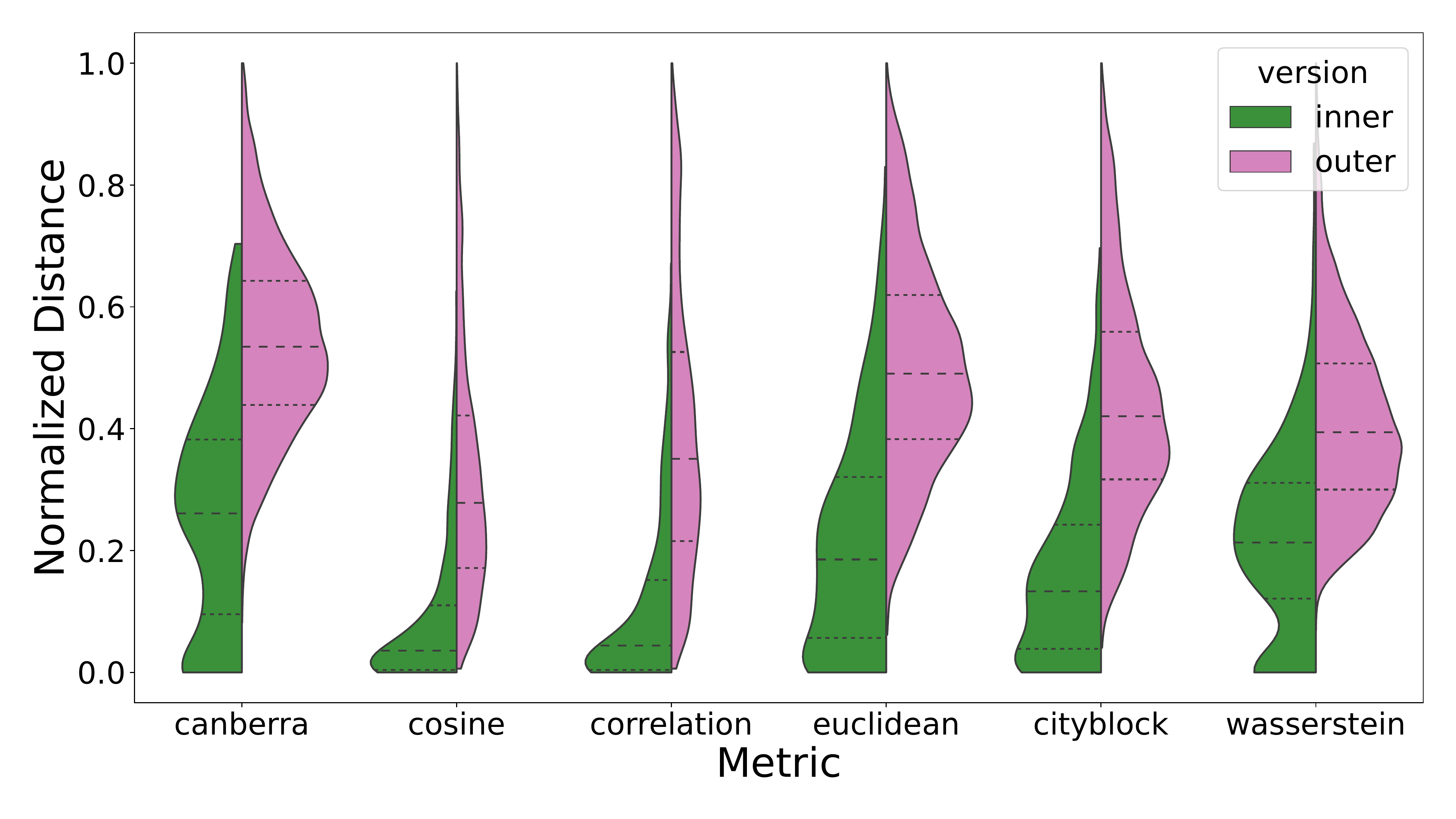}
    \caption{Distribution of normalized distances between BBOB instances of the same problem (inner) and instances of different problems (outer).}
    \label{fig:dist_inout}
\end{figure}

From Figure~\ref{fig:dist_corr}, we can see that the correlations, while clearly positive, are not perfect. 
This is especially the case when comparing the statistical distance (Wasserstein) against the vector-based distances. 
To gauge which distance metric might be preferable, we then compare the distances between instances of the \textit{same} function to the distances between instances of \textit{different} functions, as is done in Figure~\ref{fig:dist_inout}. 

Based on this comparison, we notice that the Wasserstein distance surprisingly has the lowest distinguishing power (unlike our initial intuition), while the cosine and correlation distances show a clear trend of assigning lower distances to same-function instances.

To further identify potential ways to modify the distance measures, we compare the differences in individual ELA features between same-function and different-function instances. 

\begin{figure}[!ht]
    \centering
    \includegraphics[width=0.48\textwidth,trim=10mm 10mm 5mm 9mm,clip]{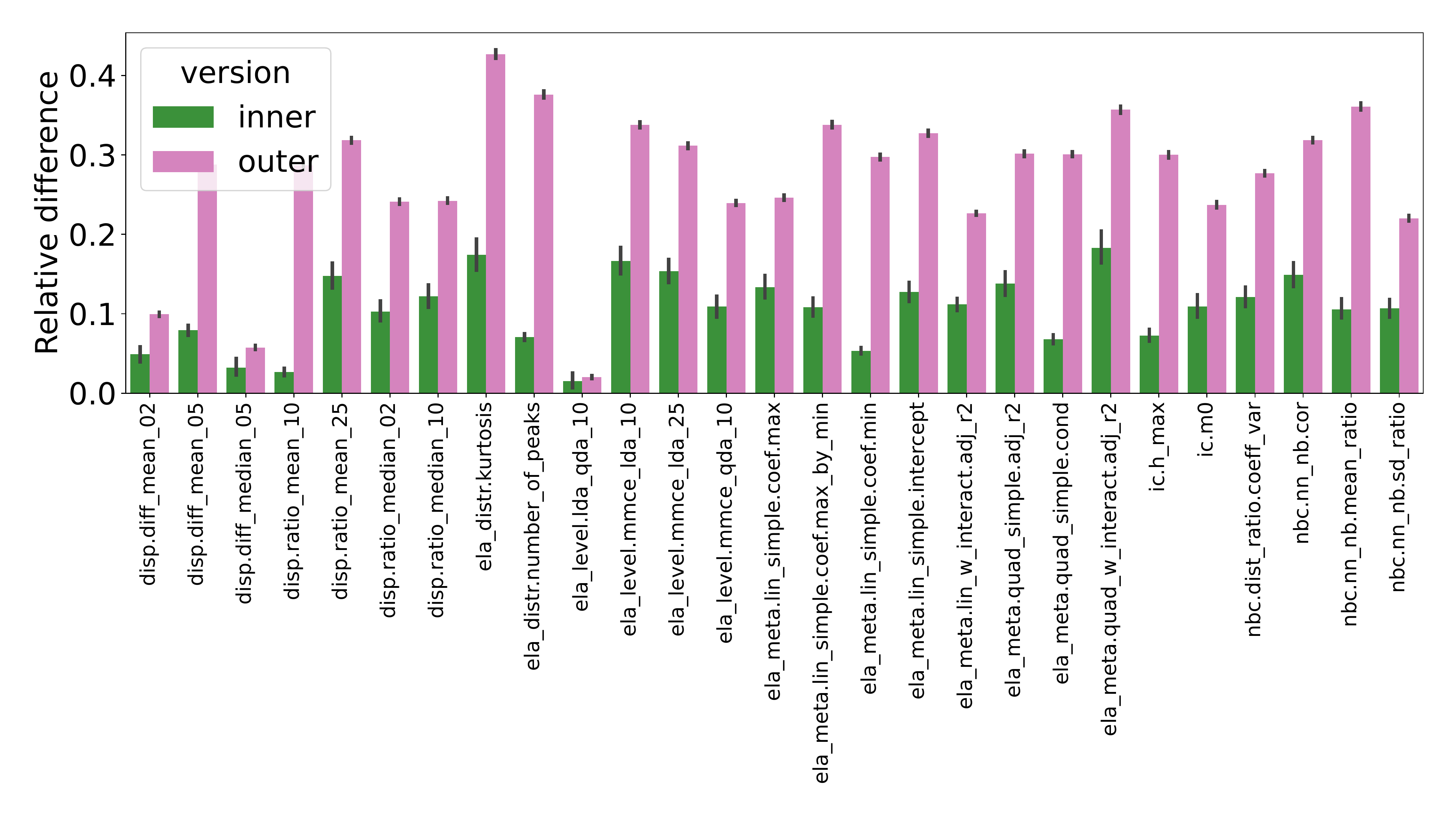}
    \caption{Relative difference (between normalized values) of each of the used ELA features, between BBOB instances of the same problem (inner) and instances of different problems (outer).}
    \label{fig:ela_diff_inout}
\end{figure}

From the results shown in Figure~\ref{fig:ela_diff_inout}, we see that some features, e.g., the conditioning of a simple quadratic model \\ (\textit{ela\_meta.quad\_simple.cond}), show very limited differences when comparing instances of the same function relative to different function instances.
As such, it is likely that these features contribute very little to any distance measure, and might be potentially removed to improve the stability, as reducing the vector dimensionality can make the distances more reliable. 

\section{Conclusions and Future Work}\label{conclusion}

In this paper, we have shown that GP can be guided by ELA features to find problems with similar high-level characteristics as a set of target problems. 
However, through this process, we highlight several potential pitfalls with this approach, illustrated by the fact that we could not accurately recreate a simple sphere problem. 

Although our results are based on a very limited set of experiments, they reveal that equal weighting of all landscape features on the distance measure leads to difficulties in focusing on the more visual high-level features. 

By comparing the differences in ELA features on the BBOB problems both between instances of the same function and instances of different functions, we show that a \textit{feature selection} mechanism should be integrated to make the fitness values more stable. 

Additionally, a \textit{weighting scheme} based on feature importances might be used to, in combination with a distance metric, more rigorously guide the GP search towards relevant function characteristics. 
Further research into ELA and other feature-free approaches are also important to improve the used approach, e.g., models such as DoE2Vec~\cite{vanstein2023doe2vec} are a starting point in this direction.

Another aspect that should be considered is the specific setting of the GP itself. 
In this work, we made use of default parameter settings with a relatively small population size for computational reasons. 
This might, however, limit the ability of GP to find diverse solutions, leading to premature convergence~\cite{schweim2021sampling}. 
Care should also be taken to include a more rigorous tree-cleaning operation, similar to \cite{tian2020arecommender}, to simplify the resulting expressions and prevent infeasible trees from being generated. 

In the future, the generation of functions with specific landscape properties has significant potential to help address real-world problems. 
By generating a diverse set of problems with similar features to a complex target problem, we can create an ideal test set for algorithm selection and configuration pipelines. 
This has clear benefits over training on a conventional surrogate, since we can avoid overfitting by considering diversified sets of problems.

\begin{acks}
The contribution of this paper was written as part of the joint project newAIDE under the consortium leadership of BMW AG with the partners Altair Engineering GmbH, divis intelligent solutions GmbH, MSC Software GmbH, Technical University of Munich, TWT GmbH. The project is supported by the Federal Ministry for Economic Affairs and Climate Action (BMWK) on the basis of a decision by the German Bundestag.

This work was performed using the ALICE compute resources provided by Leiden University.
\end{acks}

\bibliographystyle{ACM-Reference-Format}
\bibliography{main}


\end{document}